\documentclass[11pt]{article}
\usepackage[preprint]{acl}
\usepackage{times}
\usepackage{latexsym}
\usepackage[T1]{fontenc}
\usepackage[utf8]{inputenc}
\usepackage{microtype}
\usepackage{inconsolata}
\usepackage{graphicx}
\usepackage{amsmath}
\usepackage{amssymb}
\usepackage{booktabs}
\usepackage{multirow}
\usepackage{algorithm}
\usepackage{algpseudocode}
\usepackage{xcolor}
\usepackage{placeins}
\usepackage{xspace}
\usepackage{hyperref}
\usepackage{float}
\usepackage{fvextra}
\fvset{frame=single,framesep=2mm,fontsize=\footnotesize,breaklines=true,breakanywhere=true,breaksymbolleft=}
\newcommand{\methodname}{\textsc{Auto-Robotist}\xspace}
\newcommand{\evogym}{\textsc{EvoGym}\xspace}
\newcommand{\skilllib}{\mathcal{S}}

\title{When Search Becomes Memory: \\
Accelerating Robot Design Discovery with Self-Evolving Skills}
\author{
Yunfei Wang$^{*}$ \quad
Xiaohao Xu$^{*\dagger}$ \quad
Yang Li \quad
Xiaonan Huang \\
University of Michigan, Ann Arbor, MI, USA \\
\texttt{\{wyunfei,xiaohaox,yangliy,xiaonanh\}@umich.edu} \\
{\small
$^{*}$Equal contribution.
\quad
$^{\dagger}$Project lead.
}
}

\date{}
\hypersetup{
  pdftitle={When Search Becomes Memory: Accelerating Robot Design Discovery with Self-Evolving Skills},
  pdfauthor={Yunfei Wang, Xiaohao Xu, Yang Li, Xiaonan Huang},
  pdfsubject={Robot design with reusable, evidence-grounded skills},
  pdfkeywords={Auto-Robotist, robot design, large language models, skill memory, EvoGym}
}

\begin{document}
\setlength{\parskip}{0pt}
\maketitle
\begin{abstract}
Large language models (LLMs) are increasingly used as proposal generators for evolutionary robot design, yet most loops remain memoryless: simulator results shape the next population but are not preserved as reusable design knowledge. We present \methodname, a self-evolving LLM agent that distills morphology-search traces into an explicit natural-language skill library. Each skill stores a structural archetype, evidence-grounded positive and negative rules, and the evaluated designs that support them, making design memory inspectable rather than implicit in a population. During search, the agent retrieves skills to condition LLM edits of elite bodies while retaining a Genetic Algorithm (GA) mutation path for exploration; after evaluation, it updates the library through \textsc{Add}, \textsc{Diagnose}, and \textsc{Merge}. Across seven \evogym tasks spanning locomotion, traversal, and object interaction, \methodname improves cold-start
  $5{\times}5$ search and transfers learned skills to $10{\times}10$ design spaces, where reference-conditioned transfer
  outperforms or matches GA on every task when averaged across seeds. These results suggest that LLM agents can convert expensive physical evaluations into reusable, auditable design principles. Our code is publicly available at
\url{https://github.com/wangyf9/Auto-Robotist}.
\end{abstract}

\section{Introduction}
\label{sec:intro}

Robot-design search is usually judged by its final artifact: the best body found before the evaluation budget is exhausted. Yet every evaluated morphology is also an experiment about support, contact, actuation, stability, and failure. Classical morphology search uses expensive evaluations to discover robot bodies and controllers~\citep{sims2023evolving,cheney2014unshackling,cheney2018scalable,mouret2015illuminating,pugh2016quality}, but the resulting experience is typically compressed into a final population or a single elite. A human designer would not discard that evidence: after enough trials, isolated blueprints become principles, such as which frames stabilize load-bearing limbs, which symmetries suppress rotation, and which plausible structures repeatedly collapse into poor gaits.

\begin{figure}[t!]
    \centering
    \includegraphics[width=\linewidth]{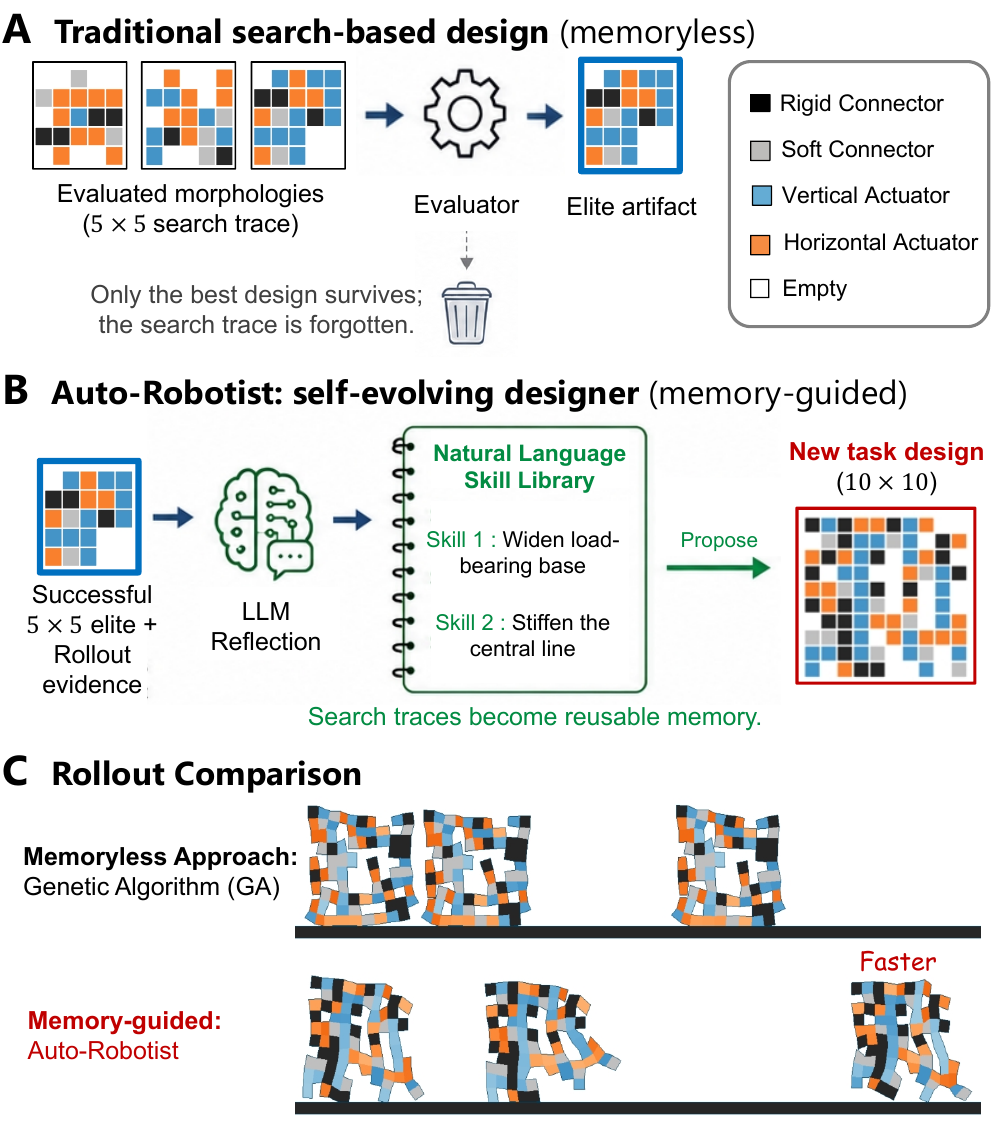}
    \caption{
    \textbf{Search traces as reusable design knowledge.}
    \textbf{(A)} Traditional search-based design is memoryless: many $5{\times}5$ morphologies are evaluated, but only the elite artifact is retained, leaving the evidence behind success and failure unused.
    \textbf{(B)} \methodname{} reflects over successful elites and rollout evidence to build a simulator-grounded natural-language skill library. Retrieved skills guide proposals for a new $10{\times}10$ design task, while each candidate remains validated by simulation.
    \textbf{(C)} In rollout comparison, memory-guided search reaches effective locomotion with fewer evaluations than a genetic algorithm baseline.
    The central distinction is that \methodname{} transfers \emph{design knowledge}, not merely robot bodies: search becomes an accumulating, auditable source of reusable  skills.
    }
    \label{fig:teaser}
\end{figure}

This paper asks whether a language agent can make the same transition: \emph{from search experience to transferable robot-design skills}. As illustrated in Figure~\ref{fig:teaser}, the key distinction is not whether an LLM can propose another morphology, but whether it can write down what costly evaluations reveal and reuse that knowledge when the design space changes. We study this question in voxel-based soft-robot morphology design in \evogym{}~\citep{bhatia2021evolution}, where an $n{\times}n$ robot is a grid of modular components and each fitness query requires Proximal Policy Optimization (PPO)~\citep{schulman2017proximal} controller training. A $10{\times}10$ robot already contains $100$ component choices, yielding an unconstrained space of $5^{100} > 10^{69}$ bodies for 5 modules. In such spaces, forgetting why a walker failed or why a carrier succeeded discards exactly the evidence that should make future search easier.

Prior work improves robot design through evolutionary algorithms, quality diversity, learned encodings, Bayesian or reinforcement-learning policies, and morphology-aware architectures~\citep{ha2019reinforcement,luck2020data,gupta2021embodied,yuan2021transform2act,lu2025bodygen,li2025freeform}. Recent LLM systems propose candidates, design rewards, or reflect on optimization outcomes~\citep{lehman2023evolution,liu2023large,ma2024eureka,song2025laser,chen2025naturalselection,ringel2025text2robot,fang2025robomore}, while NLP agents increasingly study long-term memory and reusable skills~\citep{wang2023voyager,tan2025reflective,kang2025memoryos,salama2025meminsight,xu2025amem}. Together, these lines of work raise a sharper question: \emph{what should a language agent remember after costly physical evaluations?}

A useful design memory cannot be a raw log, a copied elite body, or unconstrained reflection. It must be \textbf{abstract} enough to transfer across resolutions and tasks, \textbf{grounded} in evaluated bodies and fitness changes, and \textbf{editable} as new evidence arrives. We therefore treat memory as a revisable design theory rather than a passive transcript.

We introduce \methodname{}, a self-evolving language agent with a persistent library of robot-design skills. Each skill separates an archetype, such as \texttt{portal-frame}, from positive and negative structural rules and the observations that support them. During search, the agent retrieves relevant skills, asks the LLM to edit elite parents toward those skills, validates the proposed bodies, and evaluates them in simulation. The library is then updated through \textsc{Add}, \textsc{Diagnose}, and \textsc{Merge}: new archetypes are created from unassigned evidence, observations are distilled into rules and failure modes, and redundant skills are consolidated. The LLM weights remain fixed; the agent's explicit domain memory evolves.

\begin{figure*}[t!]
  \centering
  \includegraphics[width=\textwidth]{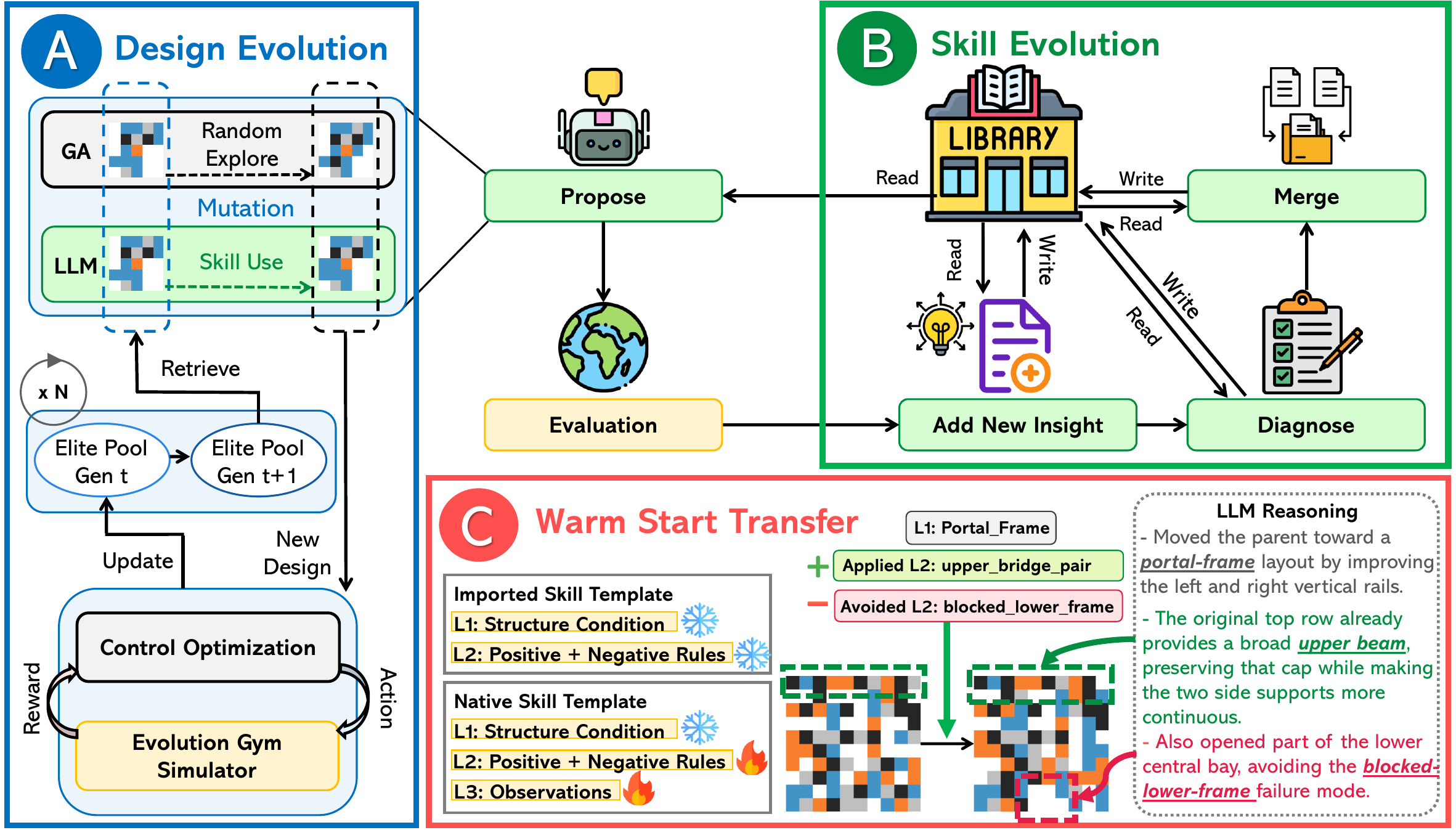}
  \caption{
    \textbf{Overview of \methodname{}.}
    \textbf{(A)} During design evolution, \methodname{} augments genetic mutation with skill-conditioned LLM proposals. Retrieved skills guide edits to elite parents, valid candidates are evaluated in \evogym{} with PPO-trained controllers, and the resulting fitness updates the elite pool.
    \textbf{(B)} During skill evolution, evaluation evidence is written back into a persistent library through \textsc{Add}, \textsc{Diagnose}, and \textsc{Merge}. This converts search traces into revisable design knowledge while keeping the LLM parameters fixed.
    \textbf{(C)} During warm-start transfer, learned skills are imported into a new design space. Each skill separates a structural archetype ($L_1$), positive and negative design rules ($L_2$), and supporting observations ($L_3$), enabling a $5{\times}5$ \texttt{portal\_frame} principle to guide a $10{\times}10$ proposal by preserving useful relations, extending support, and avoiding known failures.
  }
  \label{fig:overview}
\end{figure*}

This framing changes the evaluation standard. A proposal generator should improve one run; a design agent should leave behind knowledge that improves the next. We evaluate \methodname{} in cold-start $5{\times}5$ search, where skills must emerge online, and in $5{\times}5\!\rightarrow\!10{\times}10$ transfer, where the learned library initializes a larger design problem. Across seven \evogym{} tasks, \methodname{} exceeds GA in cold-start search and, under
  reference-conditioned transfer, improves over GA on most $10{\times}10$ tasks, especially when
  contact and load-bearing structure are central.

\vspace{2mm}
Our contributions are threefold:
\begin{itemize}
  \item We reframe LLM-based robot design as converting costly search experience into transferable design skills, rather than only generating high-fitness bodies.
  \item We introduce an evidence-grounded, self-evolving skill library that separates archetypes, rules, and observations, and couples skill-conditioned LLM proposals with simulator feedback.
  \item We evaluate on seven \evogym{} tasks and analyze the learned memory, showing competitive cold-start search, consistent $5{\times}5\!\rightarrow\!10{\times}10$ transfer gains, and evidence that skills encode reusable structural relations.
\end{itemize}
\section{Related Work}
\label{sec:related}

\paragraph{Morphology and embodiment co-design.}
Automated body-controller co-design has a long history, from evolving virtual creatures~\citep{sims2023evolving} to voxel-based soft robot evolution~\citep{cheney2014unshackling,cheney2018scalable} and quality-diversity search~\citep{mouret2015illuminating,pugh2016quality}. \evogym~\citep{bhatia2021evolution} made this setting a controlled benchmark by pairing discrete voxel bodies with reinforcement-learning controllers. Later work improved search with learned representations, Bayesian or RL-based design policies, and embodied pretraining~\citep{ha2019reinforcement,luck2020data,gupta2021embodied,yuan2021transform2act}. Recent systems further push this direction: coarse-to-fine and morphology-aware representations improve sample efficiency; BodyGen uses topology-aware self-attention and temporal credit assignment for efficient embodiment co-design~\citep{lu2025bodygen}; and freeform endoskeletal robot generation expands the design space beyond fully soft or fully rigid bodies by jointly optimizing external tissues and internal skeletons~\citep{li2025freeform}. These methods make the optimizer or representation stronger. Our question is complementary: after a design run finishes, what reusable knowledge should remain?

\paragraph{LLMs for optimization, robotics, and robot design.}
LLMs can propose candidate programs~\citep{lehman2023evolution}, optimize language objectives~\citep{liu2023large}, synthesize rewards~\citep{ma2024eureka}, and refine sim-to-real reward functions~\citep{ma2024dreureka}. For robot morphology, LASeR uses LLM-aided evolutionary search and diversity reflection~\citep{song2025laser}; other systems use foundation models to select, initialize, or co-optimize morphology from language and preferences~\citep{chen2025naturalselection,ringel2025text2robot,fang2025robomore}. These works show that LLMs are useful proposal operators. \methodname{} places the abstraction not only in the current prompt or candidate, but in a persistent skill library revised across generations and reused across runs.

\paragraph{Agent memory, reflection, and reusable skills.}
LLM agents increasingly use memory to extend beyond a single context window. Voyager stores executable skills for open-ended exploration~\citep{wang2023voyager}; reflective-memory systems revise long-term memories based on future evidence~\citep{tan2025reflective}; and recent memory architectures organize agent memories dynamically for retrieval and update~\citep{kang2025memoryos,salama2025meminsight,xu2025amem}. Our setting imposes a stricter grounding requirement: a robot-design memory is useful only if it changes future physical-search behavior and remains tied to evaluated bodies. This is why \methodname stores not only a natural-language rule but also its supporting observations and parent-relative gains.

\makeatletter
\@secpenalty=10000
\makeatother
\section{Method}
\makeatletter
\@secpenalty=-300
\makeatother
\label{sec:method}

\subsection{Problem Formulation}
\label{sec:problem}

We study morphology search for modular soft robots in \evogym{}~\citep{bhatia2021evolution}. The goal is to design a body that performs well after controller training on tasks such as walking, climbing, balancing, carrying, or pushing.

\raggedbottom
\paragraph{Design space and validity.}
At scale $n$, a robot body is an $n \times n$ matrix $d \in \mathcal{V}^{n \times n}$ whose entries are one of five voxel types: \emph{empty}, \emph{rigid connector}, \emph{soft connector}, \emph{horizontal actuator}, or \emph{vertical actuator}. Empty cells contain no material; connectors provide passive structure; actuators expand and contract along their axes. A feasible body $d\in\mathcal{D}_n\subset\mathcal{V}^{n\times n}$ must have a single connected non-empty component and at least one actuator.

\paragraph{Optimization objective.}
For task $\mathcal{T}$, each feasible body is paired with a PPO controller trained under a fixed protocol. The simulator then returns a scalar fitness $f(d,\mathcal{T})$, and morphology search solves
\begin{equation}
    d^* = \arg\max_{d \in \mathcal{D}_n} f(d,\mathcal{T}),
    \label{eq:morphology_search_objective}
\end{equation}
under a budget $B$ of body evaluations. One evaluation proposes a body, trains its controller, and measures task fitness.

\paragraph{Search trace as evidence.}
The combinatorial space is large, but the central inefficiency is epistemic: each expensive evaluation teaches something about stability, contact, actuation, symmetry, or failure, yet standard search keeps this knowledge only implicitly in the population. \methodname{} instead maintains a mutable skill library $\skilllib_t$; the LLM parameters, PPO protocol, component alphabet, and validity constraints remain fixed.

\subsection{From Search Experience to Skills}
\label{sec:method:overview}

At generation $t$, \methodname{} maintains an evaluated population $\mathcal{P}_t$ and a skill library $\skilllib_t$. Algorithm~\ref{alg:autorobotist} alternates between \emph{using} memory and \emph{updating} memory. First, the agent retrieves task- and scale-relevant skills, samples one by evidence-weighted utility, and asks the LLM to edit elite parents toward the retrieved rule; a GA mutation path runs in parallel for coverage and invalid-output fallback. Then, evaluated children become observations, are routed to existing skills or an unassigned evidence pool, and are distilled into revised rules. The loop therefore compresses individual simulations into reusable design knowledge.

\begin{algorithm}[t]
\small
\caption{\methodname: converting search traces into transferable skills}
\label{alg:autorobotist}
\begin{algorithmic}[1]
\State Initialize evaluated population $\mathcal{P}_0$ and skill library $\skilllib_0$.
\For{generation $t=0,\ldots,T-1$}
  \State Retrieve task- and scale-relevant skills from $\skilllib_t$.
  \State Propose LLM children by editing elite parents toward retrieved skills.
  \State Propose GA children by random mutation for coverage.
  \State Reject invalid children; repair or replace them when possible.
  \State Evaluate valid children with PPO to obtain fitness and parent-relative gain.
  \State Route each observation to an existing skill or to the unassigned evidence pool.
  \State Update $\skilllib_{t+1}$ with \textsc{Add}, \textsc{Diagnose}, and \textsc{Merge}.
\EndFor
\State \Return best evaluated body and final skill library.
\end{algorithmic}
\end{algorithm}

Lines 3--5 of Algorithm~\ref{alg:autorobotist} show how the library steers exploration; Lines 8--9 show how simulator evidence rewrites the library. This separation prevents the LLM from becoming an oracle: its proposals must pass validity checks, its abstractions are trusted only after evaluation, and its memory is tied to observed fitness changes.

\subsection{Skill Representation}
\label{sec:method:skill_lib}

A skill $s\in\skilllib$ is a three-level record:
\begin{itemize}
  \setlength{\itemsep}{0pt}
  \item $L_1$ is the \textbf{archetype}: a compact structural concept with an applicability condition, e.g., \texttt{portal-frame} for load-bearing arches around actuators.
  \item $L_2$ contains \textbf{rules}: positive rules that tend to improve fitness and negative rules that describe failure modes. Each rule stores a structural description, supporting evidence, and running mean parent-relative gain.
  \item $L_3$ contains \textbf{observations}: evaluated child and parent bodies, task, scale, fitness, gain, validity status, and proposal attribution.
\end{itemize}
This structure gives memory stability and accountability: $L_1$ is retrievable, $L_2$ specifies how to instantiate or avoid the concept, and $L_3$ preserves the audit trail. Rules use morphology-level relations rather than absolute coordinates; ``support a central actuator column with a rigid side frame'' can transfer, while ``set cell $(4,2)$ to rigid'' cannot.

\subsection{Skill-Conditioned Proposal}
\label{sec:method:proposal}

Let $g_{s,i}$ be the parent-relative fitness gain of the $i$-th observation attributed to skill $s$, and let $n_s$ be the number of such observations. We score each skill by a smoothed usefulness estimate:
\begin{equation}
  w_s = \frac{1 + \sum_{i=1}^{n_s} \mathrm{clip}(g_{s,i}/\delta_{\max}, 0, 1)}{2+n_s}.
  \label{eq:skill_weight}
\end{equation}
After normalizing these scores over retrieved skills, the agent samples a skill and prompts the LLM with the task, scale, parent body, archetype, relevant rules, and a strict output schema. Positive evidence increases reuse probability; repeated non-improvement lets better-supported skills dominate. The child must satisfy legal-voxel, connectivity, and actuator-presence checks before evaluation. Local, unambiguous invalidities are repaired; otherwise, the slot is filled by GA mutation.

\subsection{Library Maintenance}
\label{sec:method:operations}

After each evaluated batch, the agent updates the library with three operations.

\paragraph{\textsc{Add}.}
Observations that do not match existing skills enter an unassigned evidence pool. Once per generation, the LLM receives this evidence and the current archetypes, and may create a new $L_1$ only for a recurring structure not already represented. This prevents one-skill-per-anecdote growth.

\paragraph{\textsc{Diagnose}.}
For each skill receiving new evidence, the agent updates its rules. If a child targeted a rule, its gain updates that rule's running mean:
\begin{equation}
  \bar g_\ell \leftarrow \bar g_\ell + \frac{g_i-\bar g_\ell}{m_\ell},
  \label{eq:rule_running_mean}
\end{equation}
where $m_\ell$ is the updated support count. Rule-free observations are passed to the LLM, which decides whether they support, contradict, create, or leave undecided a rule. Negative rules are first-class entries because failure modes often transfer as reliably as successes.

\paragraph{\textsc{Merge}.}
LLM-generated archetypes can be redundant. The agent clusters skills by $L_1$ identity and task family, merging only those that describe the same structural principle. Rules and observations from absorbed skills are preserved, consolidating evidence without erasing the audit trail.

\subsection{Cross-Scale Transfer}
\label{sec:method:transfer}

To transfer from source scale to target scale, the target run imports the source skill library. Imported skills retain $L_1$ and $L_2$ but drop raw source observations, so they act as priors rather than target-scale evidence. During the first target generations, retrieval emphasizes imported skills; later observations enter the unassigned pool and can form native target-scale skills. The agent thus moves from prior-guided exploration to target-grounded refinement.

No voxel upsampling is required. A $5{\times}5$ rule such as ``use a rigid side frame to stabilize vertical actuators'' can guide $10{\times}10$ search because it names relations among functional parts. This is the core advantage of representing search experience as design skills rather than  raw examples.

\setlength{\parskip}{0pt plus 1pt}
\flushbottom
\section{Experiments}
\label{sec:experiments}

A design-memory agent should pass three tests: improve the run that creates the memory, reuse that memory after the design space changes, and transfer physical relations rather than copy an elite body. We therefore ask: (i) can skills emerge from an empty library during $5{\times}5$ search; (ii) do $5{\times}5$ skills help in the larger $10{\times}10$ space; and (iii) are gains better explained by relational rules than by direct resizing or visual imitation? We answer these questions with fitness curves, endpoint summaries, rollouts, and library inspection.

\subsection{Experimental Setup}
\label{sec:exp:setup}

\paragraph{Benchmark.}
We evaluate on \evogym~\citep{bhatia2021evolution}, a 2D soft-robot benchmark where performance depends on support, contact, actuation, and task geometry. We use seven tasks spanning locomotion, traversal, balance, and object interaction: Walker, BridgeWalker, Balancer, Carrier, Climber, Jumper, and Pusher. Each task is tested in cold-start $5{\times}5$ search and $5{\times}5\rightarrow10{\times}10$ transfer.

\paragraph{Budget and controller training.}
At $5{\times}5$, Walker and BridgeWalker use $250$ morphology evaluations; Balancer, Carrier, and Climber use $500$; and Jumper and Pusher use $750$. At $10{\times}10$, all tasks use $1{,}000$ evaluations. Each body is evaluated by training a PPO controller~\citep{schulman2017proximal} for $5.12{\times}10^{5}$ environment steps, following \evogym{}~\citep{bhatia2021evolution}. Full hyperparameters are in Appendix~\ref{app:exp:solo_hyperparams}.

\paragraph{Baselines and variants.}
The main baseline is the standard \evogym{} genetic algorithm (GA), run with the same budget, population size $25$, and survivor-rate schedule. For transfer, \emph{with reference} imports the $5{\times}5$ library and shows the source elite in the prompt; \emph{skill-only} imports the library but removes the source-body exemplar, isolating rule transfer from visual imitation.

\paragraph{Metrics and reporting.}
Let $f_A(e)$ and $f_G(e)$ denote the best fitness found by \methodname and GA within the first $e$ morphology evaluations, under total budget $B$. We report best-fitness-so-far curves and three scalar summaries. \emph{Maximal Fitness} is the final value $f_A(B)$. \emph{Convergence Speedup (S)} compares how quickly \methodname reaches the GA endpoint with how quickly GA reaches its own endpoint,
\begin{equation}
  S = \frac{\min\{e : f_G(e) \geq f_G(B)\}}{\min\{e : f_A(e) \geq f_G(B)\}},
  \label{eq:speedup}
\end{equation}
with $S>1$ indicating that \methodname reaches the GA endpoint faster than GA does. \emph{Lead Fraction (L)} is the proportion of the search budget for which \methodname is strictly ahead,
\begin{equation}
  L = \bigl|\{e : f_A(e) > f_G(e)\}\bigr| / B.
  \label{eq:lead_fraction}
\end{equation}
All curves plot cumulative morphology evaluations, so methods that consume the budget at different rates are compared on equal footing. We treat the results as descriptive comparisons under a fixed evaluation protocol.

\begin{table}[t]
  \centering
  \small
  \caption{
    \textbf{$5{\times}5$ cold-start skill discovery.}
    $\Delta=\textsc{Auto-Robotist}-\mathrm{GA}$.
    $S$ is convergence speedup relative to GA, and $L$ is lead fraction over the budget.
    \textsc{Auto-Robotist} reports mean $\pm$ s.d.\ over three seeds ($0$, $9$, and $42$).
  }
  \vspace{-2mm}
  \label{tab:exp2_5x5}
  \resizebox{0.48\textwidth}{!}{
  \begin{tabular}{lccccc}
  \toprule
  \textbf{Task}
  & \textbf{GA}
  & \textbf{\textsc{Auto-Robotist}}
  & $\boldsymbol{\Delta}$
  & $\boldsymbol{S}$
  & $\boldsymbol{L}$ \\
  \midrule
  Walker
  & 9.54
  & \textbf{9.55}{\scriptsize$\pm$0.01}
  & $+0.01$
  & 1.30
  & 0.67 \\

  BridgeWalker
  & 3.65
  & \textbf{3.69}{\scriptsize$\pm$0.05}
  & $+0.04$
  & 1.28
  & 0.53 \\

  Balancer
  & 0.13
  & \textbf{0.15}{\scriptsize$\pm$0.01}
  & $+0.02$
  & 1.13
  & 0.45 \\

  Carrier
  & 6.40
  & \textbf{8.19}{\scriptsize$\pm$1.21}
  & $+1.79$
  & 2.76
  & 0.87 \\

  Climber
  & 0.57
  & \textbf{0.61}{\scriptsize$\pm$0.06}
  & $+0.04$
  & 1.24
  & 0.65 \\

  Jumper
  & 5.58
  & \textbf{6.67}{\scriptsize$\pm$0.33}
  & $+1.09$
  & 2.16
  & 0.83 \\

  Pusher
  & 8.45
  & \textbf{8.92}{\scriptsize$\pm$1.36}
  & $+0.47$
  & 1.74
  & 0.71 \\
  \bottomrule
  \end{tabular}
  }
  \end{table}

\subsection{Implementation Details}
\label{sec:method:impl}

For \methodname{}, \texttt{gpt-5.5} is used for \textsc{Propose}, \textsc{Add}, \textsc{Diagnose}, and \textsc{Merge}, with temperature $1.0$. Each generation evaluates $25$ designs: $15$ skill-conditioned Path~A proposals and $10$ GA-style Path~B proposals. Thus, the agent biases search toward evidence-supported hypotheses without removing mutation-based exploration. Candidate bodies are checked for legal voxel types, connectivity, and actuator presence before simulation; invalid or unrecoverable LLM outputs are replaced by GA mutations.

\subsection{Cold-Start Discovery: Can Skills Emerge from Search?}
\label{sec:exp:5x5}

\begin{figure*}[t]
  \centering
  \includegraphics[width=\textwidth]{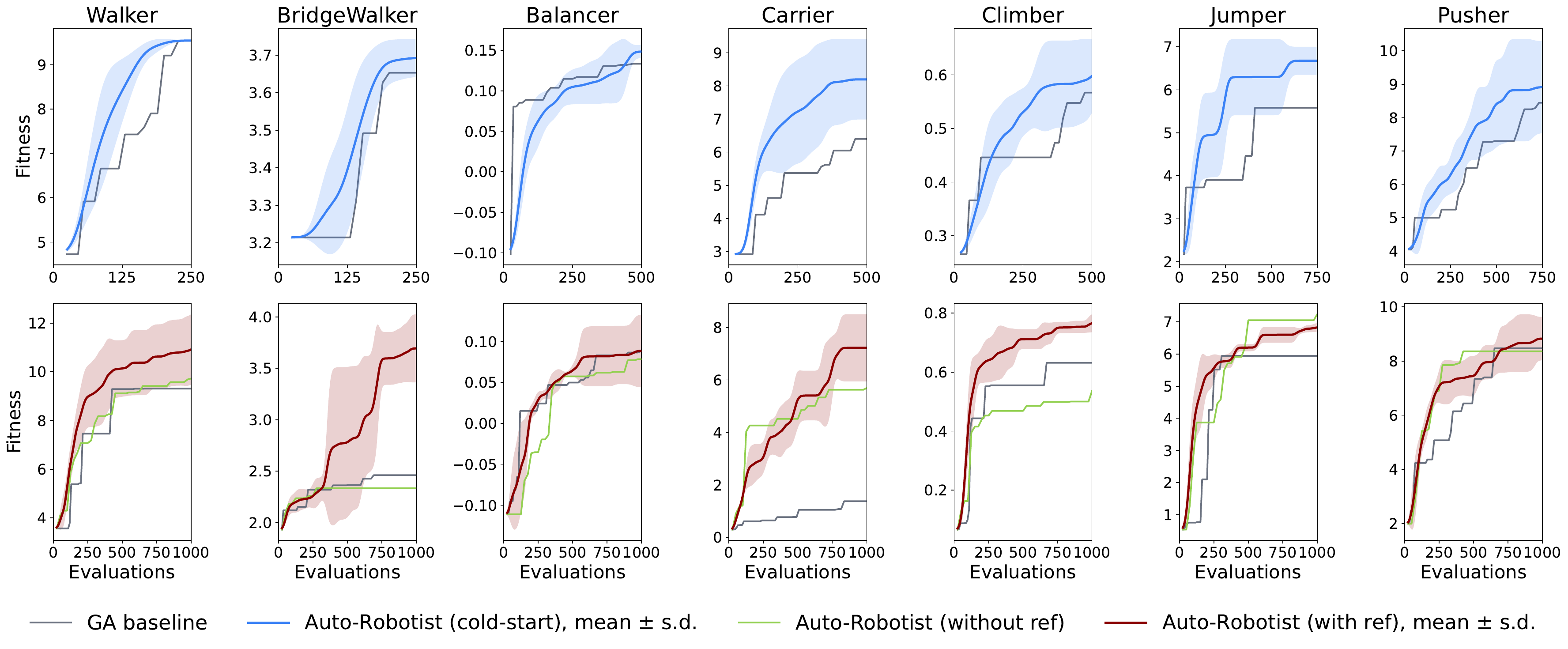}\vspace{-3mm}
\caption{
  \textbf{Best fitness vs. morphology evaluations.} \textbf{Top}: $5{\times}5$ cold-start search, where \methodname{} learns from empty memory. \textbf{Bottom}: $5{\times}5\rightarrow10{\times}10$ transfer, where the learned library is reused in a larger space. The green curve removes the source-body exemplar, isolating skill transfer from visual imitation.
}\label{fig:main_results}\vspace{0mm}
\end{figure*}

Cold-start search tests whether useful memory can be built during the same run that uses it. As shown in
Table~\ref{tab:exp2_5x5} and the top row of Figure~\ref{fig:main_results}, \methodname{} on average
exceeds GA on all seven tasks, with $1.66\times$ average convergence speedup and a $67\%$ average
lead fraction. The strongest endpoint gains appear on Carrier, Jumper, and Pusher, where success depends on
persistent contact, load paths, and coordinated impulse rather than merely adding actuation. Thus, the library primarily improves the search trajectory, especially early exploration, while GA remains a strong optimizer for local search and exploration.

\begin{figure}[t]
  \centering
  \includegraphics[width=0.48\textwidth]{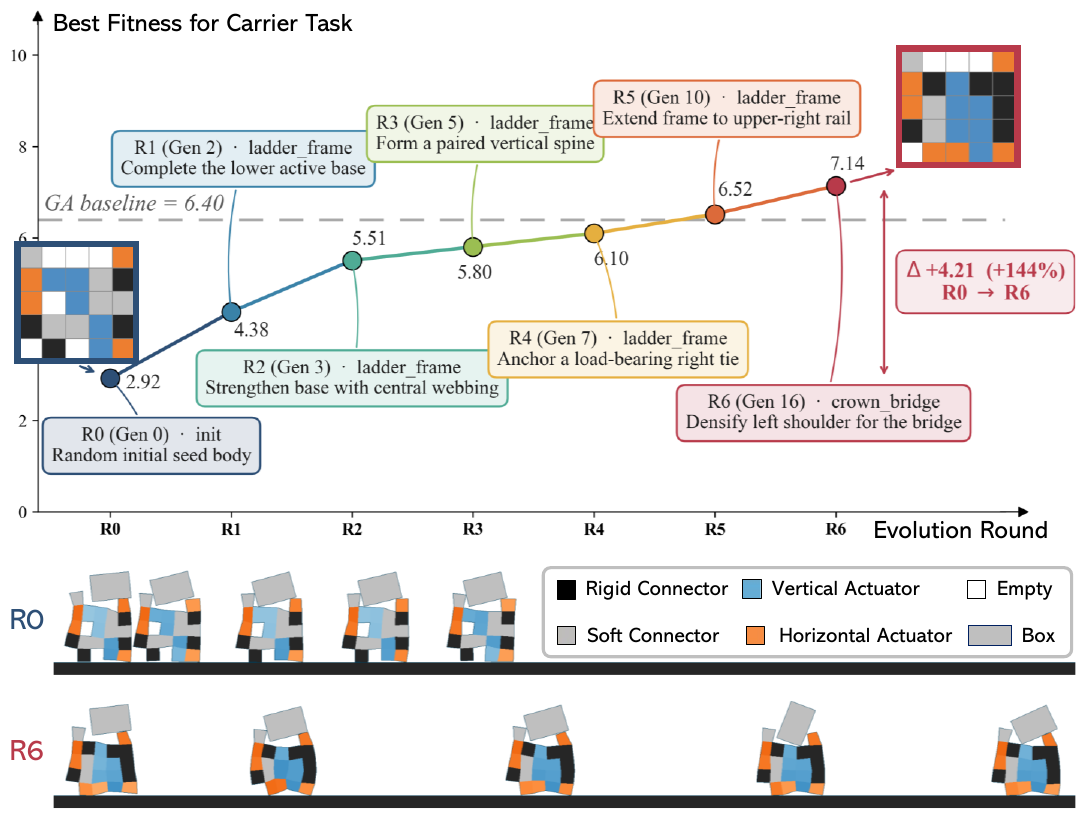}\vspace{-1mm}
  \caption{
    \textbf{Design evolution trace of \textsc{Auto-Robotist} on Carrier.} Each point marks the best body after an evidence-guided library update, annotated with the skill or rule that motivated the next design edit. \methodname{} transforms a weak random seed into a load-bearing carrier by completing the active base, adding central webbing, forming a vertical spine, and densifying bridge support. The bottom rollouts show that the final body better preserves payload support over time.
  }
  \label{fig:carrier_evolution_trace}\vspace{-2mm}
\end{figure}

\paragraph{Design evolution trace.}
Figure~\ref{fig:carrier_evolution_trace} shows how memory is formed during a single Carrier run. Starting from a random seed body, the agent repeatedly writes evaluation evidence into named structural hypotheses, such as completing a lower active base, strengthening central webbing, forming a paired vertical spine, and densifying a left shoulder for bridge-like support. The final body improves from $2.92$ to $7.14$, surpassing the GA endpoint of $6.40$. This trace makes the cold-start result concrete: the agent is not only sampling new bodies, but converting evaluated edits into reusable rules about payload contact and load-bearing structure.

\begin{table*}[!t]
  \centering
  \small
  \caption{\textbf{$10{\times}10$ cross-scale transfer.}
  \methodname{} vs the Genetic Algorithm (GA) baseline at $10{\times}10$.
  \emph{Upsampling} columns evaluate the $5{\times}5$ elite after upsampling to $10{\times}10$ via $2{\times}2$ voxel tiling;
  \emph{Full search} columns use a $1{,}000$-eval $10{\times}10$ budget.
  \emph{w/ ref} and \emph{w/o ref} indicate whether a $5{\times}5$ reference design is included in the proposal prompt. 
  \emph{w/ ref} reports mean $\pm$ s.d.\ over three seeds (0, 9, 42).
  $\Delta = \text{\textsc{Auto-Robotist} w/ ref} - \text{GA}$;
  $S$ and $L$ are computed for \methodname{} \emph{w/ ref} against GA.}
  \label{tab:exp5}
  \vspace{-2mm}

  \setlength{\tabcolsep}{4pt}
  \begin{tabular}{lcccccccc}
  \toprule
  \textbf{Task}
  & \multicolumn{2}{c}{\textit{Upsampling $5{\times}5$ to $10{\times}10$}}
  & \multicolumn{3}{c}{\textit{Full search at $10{\times}10$}}
  & \multicolumn{3}{c}{\textit{w/ ref vs.\ GA}} \\
  \cmidrule(lr){2-3}
  \cmidrule(lr){4-6}
  \cmidrule(lr){7-9}
  & GA
  & Auto-Robotist
  & GA
  & Auto-Robotist w/o ref
  & Auto-Robotist w/ ref
  & $\Delta$
  & $S$
  & $L$ \\
  \midrule
  Walker
  & 4.67 & 4.20 & 9.31 & 9.72
  & \textbf{10.98}{\scriptsize$\pm1.45$}
  & $+1.67$ & 3.27 & 0.75 \\

  BridgeWalker
  & 1.24 & 0.42 & 2.46 & 2.33
  & \textbf{3.69}{\scriptsize$\pm0.33$}
  & $+1.23$ & 1.47 & 0.66 \\

  Balancer
  & -1.44 & -0.14 & \textbf{0.09} & 0.08
  & \textbf{0.09}{\scriptsize$\pm0.04$}
  & $+0.00$ & 1.81 & 0.52 \\

  Carrier
  & 3.41 & 0.16 & 1.38 & 5.69
  & \textbf{7.22}{\scriptsize$\pm1.28$}
  & $+5.84$ & 8.36 & 0.95 \\

  Climber
  & 0.34 & 0.29 & 0.63 & 0.54
  & \textbf{0.77}{\scriptsize$\pm0.03$}
  & $+0.13$ & 3.11 & 0.93 \\

  Jumper
  & 1.55 & 2.71 & 5.94 & \textbf{7.24}
  & 6.83{\scriptsize$\pm0.14$}
  & $+0.89$ & 1.19 & 0.87 \\

  Pusher
  & 5.85 & 0.76 & 8.47 & 8.36
  & \textbf{8.83}{\scriptsize$\pm0.80$}
  & $+0.36$ & 0.97 & 0.74 \\
  \bottomrule
  \end{tabular}
  \vspace{-2mm}
  \end{table*}

\begin{figure*}[t]
  \centering
  \includegraphics[width=\textwidth]{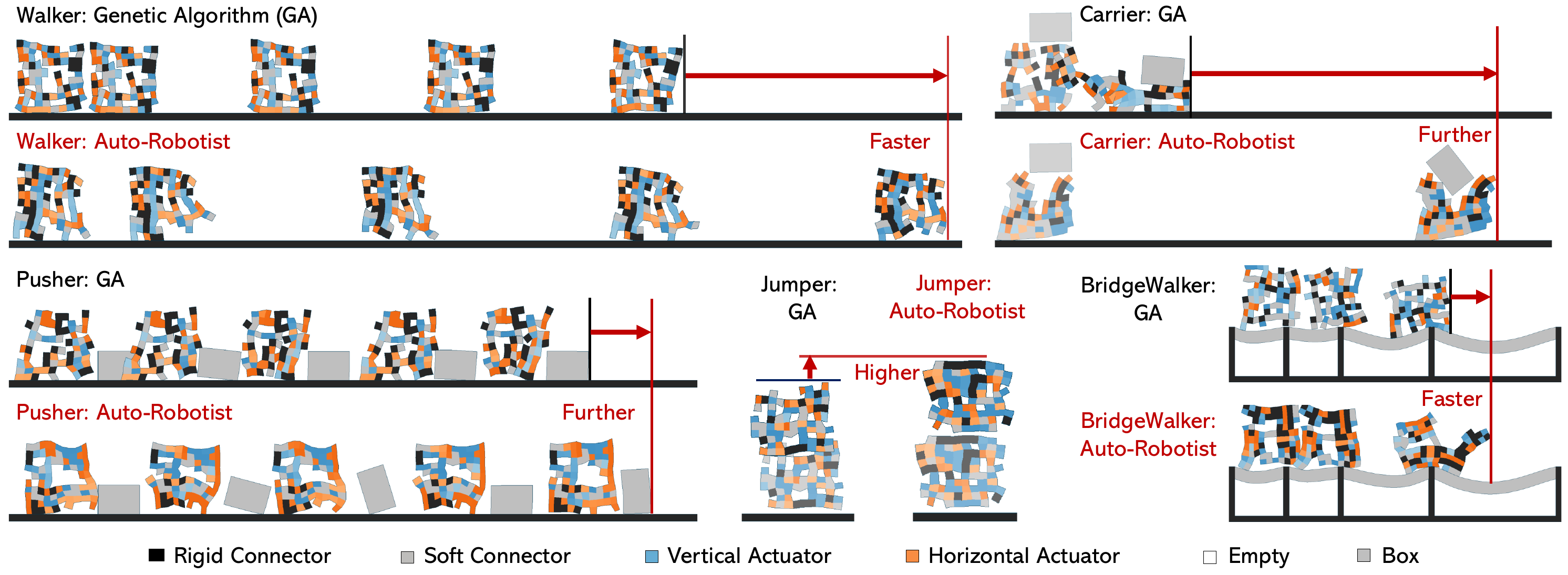} \vspace{-6mm}
  \caption{
    \textbf{Representative best $10{\times}10$ transfer rollouts.} \methodname{} more often preserves support and contact through time, yielding faster movement, higher jumps, or farther object displacement than GA.
  }
  \label{fig:transfer_rollouts_compact}\vspace{-0mm}
\end{figure*}

\raggedbottom
\subsection{Cross-Scale Transfer: Do Skills Survive a Larger Design Space?}
\label{sec:exp:10x10}

The bottom row of Figure~\ref{fig:main_results} and Table~\ref{tab:exp5} show that $5{\times}5$ skills remain useful
at $10{\times}10$. Reference-conditioned \methodname{} on average exceeds GA on six of seven tasks
and ties on the seventh, improving endpoint fitness by $+1.45$. The largest gain is Carrier ($+5.84$), where success
requires moving while preserving payload support; Walker and BridgeWalker also benefit, indicating that frames, legs,
and compliant contact pads survive scale change. Pusher is the only speedup exception ($S=0.97$): \methodname{}
finishes better but crosses the GA endpoint slightly later, suggesting a quality-over-earliness tradeoff rather than
failed transfer.
\flushbottom

\paragraph{Rollout evidence.}
Figure~\ref{fig:transfer_rollouts_compact} explains the aggregate gains behaviorally. GA elites often contain enough material and actuation to move, but lose the task-critical relation: contact breaks, support bends, or energy is spent on internal deformation. \methodname{} more often preserves the functional motif--a stable walking frame, a supported carrier body, or a pusher that keeps object contact. The learned memory therefore changes which physical invariants search protects, not only the final fitness score.

\begin{figure}[t]
  \centering
  \includegraphics[width=0.48\textwidth]{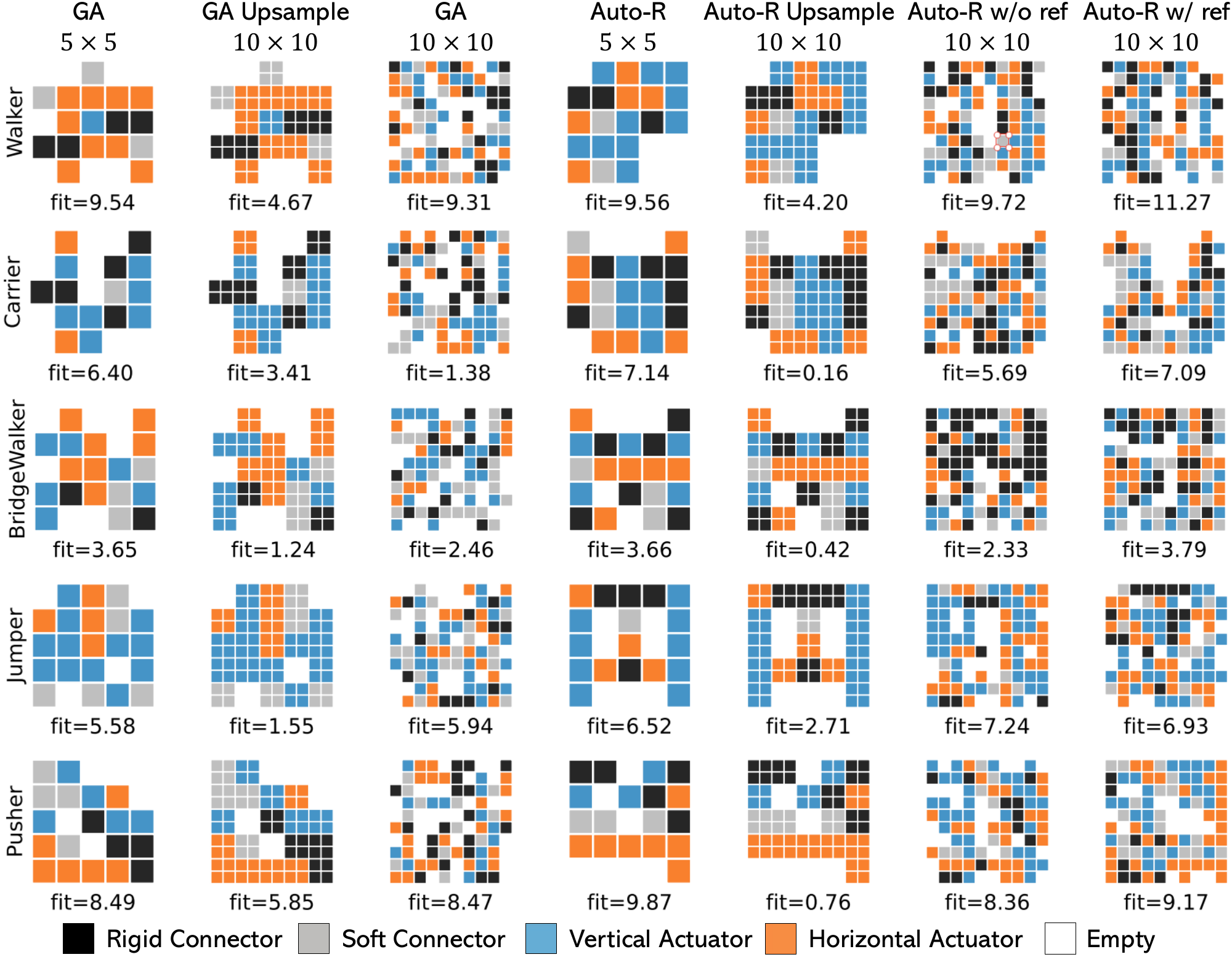}\vspace{0mm}
  \caption{\textbf{Elite robot designs across $5{\times}5$ and $10{\times}10$.} Direct upsampling preserves geometry but often loses function; skill-guided search reconfigures source motifs into viable target-scale morphologies.
  } \vspace{0mm}
  \label{fig:exp5_designs_grid}
\end{figure}

\paragraph{Why not just upsample the source robot morphology?}
Table~\ref{tab:exp5} and Figure~\ref{fig:exp5_designs_grid} give a negative control. Deterministic $2{\times}2$ tiling preserves source pixels but often destroys source function, especially for Carrier and Pusher, where scaling changes contact, leverage, and load distribution. Full $10{\times}10$ search is necessary; \methodname{} helps because it transfers morphological relations--support around actuators, compliant contact backed by structure, balanced limb placement--that \mbox{target-scale} search can re-instantiate.

\paragraph{Skill memory versus design copying.}
Figure~\ref{fig:exp5_designs_grid} also separates rule transfer from exemplar copying. Skill-only transfer removes the source-body exemplar but retains the library; it still beats GA on Walker, Carrier, and Jumper and remains competitive on Pusher. Thus, the library stores functional information beyond the appearance of a $5{\times}5$ elite. BridgeWalker, Balancer, and Climber benefit most when the reference body is also supplied, indicating that some abstract rules need a geometric anchor. The best transfer combines both: skills specify \emph{what relation should hold}, while the reference suggests \emph{where to begin instantiating it}.

\subsection{Anatomy of a Learned Skill Library}
\label{sec:exp:library}

\begin{figure}[t]
  \centering
  \includegraphics[width=0.98\columnwidth]{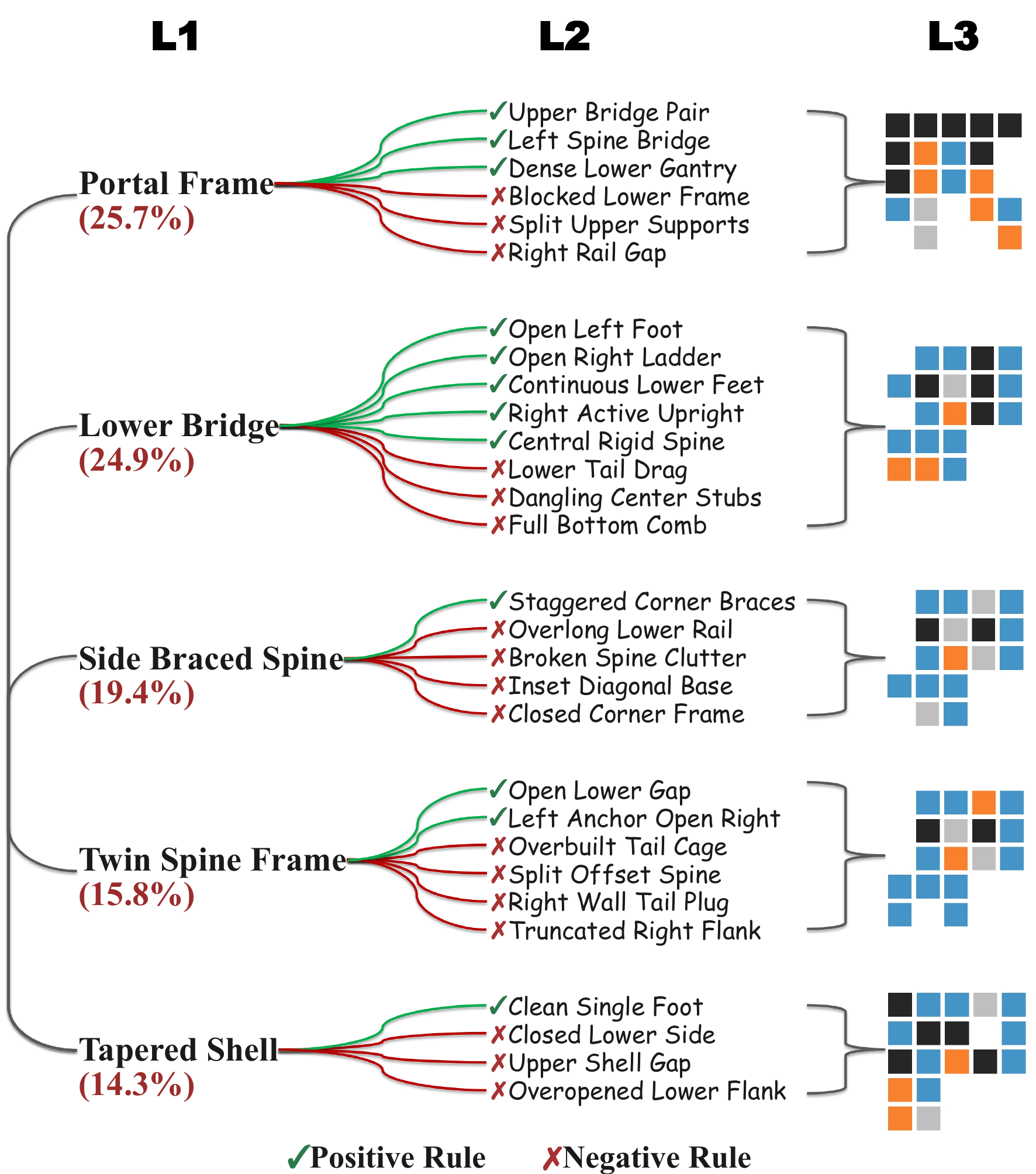} \vspace{0mm}
  \caption{
    \textbf{A learned Walker skill library.} Evaluations are distilled into $L_1$ archetypes, $L_2$ positive/negative rules, and $L_3$ supporting observations, exposing both reusable structures and grounded failure modes.
  }\vspace{0mm}
  \label{fig:walker_library}
\end{figure}

Figure~\ref{fig:walker_library} inspects the Walker $5{\times}5$ library. The memory stabilizes at five $L_1$ archetypes--lower-bridge, portal-frame, twin-spine-frame, side-braced-spine, and tapered-shell--with $12$ positive and $17$ negative $L_2$ rules. Sampling weights remain distributed across all five skills, suggesting that \textsc{Merge} reduces redundancy without collapsing the library into one template.

The library's value is abstraction with evidence. Portal-frame and lower-bridge skills encode support paths that can be re-instantiated at another scale; negative rules preserve recurring traps such as brittle bottom connections, overlong rails, and closed-shell gaps. Failure evidence becomes a constraint on future proposals. Transfer is therefore not coordinate copying, but skill-guided re-search using compact hypotheses about load paths, contact surfaces, and actuator placement.

\section{Conclusion}
\label{sec:conclusion}

We introduced \methodname{}, a self-evolving language agent that turns robot morphology search from disposable trial-and-error into transferable design knowledge. Rather than retaining only the best body, the agent distills evaluated designs into an explicit skill library of archetypes, evidence-grounded rules, and supporting observations. Across \evogym{} tasks, these skills emerge during cold-start $5{\times}5$ search and improve $10{\times}10$ transfer, especially when contact, support, and load-bearing structure matter. More broadly, \methodname{} suggests a path for improving LLM agents in expensive scientific and engineering domains: not only generating better proposals, but learning what to remember. 

\section{Limitations}
\label{sec:limitations}

Our experiments study 2D voxel morphology search in \evogym{} under fixed-budget PPO evaluation. This controlled setting lets us isolate whether search traces can be distilled into transferable design skills, but it does not by itself establish performance for 3D robots, hardware deployment, or environments with unmodeled dynamics. The current skill library is also grounded in simulator feedback from a fixed component alphabet, controller-training protocol, and validity checker; transferring the approach to other morphologies or physical platforms will require new safety constraints, uncertainty-aware skill scoring, and sim-to-real validation. The main practical risk is over-interpreting simulation-derived skills as hardware-ready designs. We therefore view \methodname{} as a research framework for auditable design-memory construction, not an autonomous hardware-deployment pipeline. Future work should extend the framework to 3D and hardware-facing morphologies, open-source LLM backbones, explicit rule-conflict resolution, calibrated uncertainty estimates, and stress tests for unsafe or brittle designs. These limitations do not change the central thesis: expensive physical evaluations should leave grounded abstractions that future agents can reuse, inspect, and audit.
\bibliography{custom}
\appendix
\clearpage
\section{Prompt Templates}
\label{app:prompts}

This appendix lists the prompt templates used by \methodname{}. Curly-brace fields such as \texttt{\{task\_desc\}} are runtime substitutions; all other text is sent to the LLM as is. Sections follow the loop steps of \S\ref{sec:method}.

\subsection{Shared Blocks}
\label{app:prompts:shared}

\paragraph{Voxel legend.}

\begin{Verbatim}
Voxel type integers (robot body):
  0 = EMPTY   (no robot voxel)
  1 = RIGID   (structural, cannot actuate)
  2 = SOFT    (deformable, passive)
  3 = H_ACT   (horizontal actuator -- expands/contracts horizontally)
  4 = V_ACT   (vertical actuator -- expands/contracts vertically)
  5 = FIXED   (environment geometry only; do not use in robot bodies)
\end{Verbatim}

\paragraph{Body requirements.}

\begin{Verbatim}
- body is a {grid_size} x {grid_size} integer grid (rows top-to-bottom, columns left-to-right)
- robot-body entries must be 0, 1, 2, 3, or 4
- the robot MUST be fully connected (all non-empty voxels reachable via 4-connectivity, no isolated groups)
- MUST contain at least one actuator (3=H_ACT or 4=V_ACT)
- do not use 5=FIXED in robot bodies
\end{Verbatim}

\subsection{Propose}
\label{app:prompts:propose}

\paragraph{Cold-start initialization.}

\begin{Verbatim}
You are an expert soft-robot morphology designer for the EvoGym simulation platform.

Task: {task_desc}

Voxel grid legend:
{voxel_legend}

Generate exactly {n_designs} diverse robot designs as a JSON object with this schema:
{
  "designs": [
    {
      "body": <{grid_size}x{grid_size} integer matrix>,
      "reasoning": "brief explanation of design choices"
    }
  ]
}

Requirements:
<BODY_REQUIREMENTS>
- Explore diverse structures -- vary leg count, body shape, actuator placement, symmetry
- Every design must be structurally different from the others
\end{Verbatim}

\paragraph{Skill-conditioned mutation.}

\begin{Verbatim}
You are an expert soft-robot morphology designer for the EvoGym simulation platform.

Task: {task_desc}

{transfer_context_block}
Voxel grid legend:
{voxel_legend}

Here is the parent design to mutate (fitness={parent_fitness:.3f}):
{parent_body}

Skill assignments for this proposal batch:
{skill_assignments_block}

{static_reference_block}
Previous mutation history on this exact parent:
{history_block}

Your task:
Propose exactly {n_designs} new mutations of this parent, one for each assigned slot above,
using a two-part process:

1. Direction from the assigned skill:
   Use the skill's L1 condition as the target structural archetype for that slot.
   If the skill has no L2 rules yet, still move the parent toward the L1 condition.

2. Tactics from L2 rules and exact-parent history:
   Use L2 positive rules as helpful sub-patterns when they fit this parent.
   Avoid L2 negative rules when relevant.
   Use exact-parent history to avoid repeats and avoid edits that already failed on this parent.

Each history entry gives you raw evidence only:
- the exact child_fitness achieved on this parent
- the voxel_diff from parent to child
- the full child_body after that mutation

Use this evidence directly to judge which edits seem promising, harmful, or ambiguous.

Hard constraints:
- You must produce new child bodies distinct from all history entries listed above.
- For each slot, use the specific assigned skill for that slot only.
- For each slot, also output `intended_leaf_id`: the L2 leaf you are trying to instantiate.
  IMPORTANT: this MUST be the leaf_id (e.g. "pos_0", "neg_1") shown in the skill's L2 rules block, NOT the claim string. Look for "leaf_id=..." in the rules listing.
  If the assigned skill has no leaves yet, output null.
- If an assigned skill is not a natural fit for this parent, propose the smallest edit that still moves in that skill direction without destroying the parent's core structure.
- Do not refuse to propose. An imperfect mutation is still useful learning material.

Generate exactly {n_designs} mutated variations of this parent as a JSON object with this schema:
{
  "designs": [
    {
      "slot_index": 0,
      "body": <{grid_size}x{grid_size} integer matrix>,
      "reasoning": "what you changed from the parent and why",
      "based_on_skill": "skill_id string or null",
      "intended_leaf_id": "leaf_id string or null"
    }
  ]
}

Requirements:
<BODY_REQUIREMENTS>
- Each design should modify {mutation_range} voxels from the parent -- keep what works, change what could improve
- L1 chooses the mutation direction; L2 rules and exact-parent history choose the concrete edits
- History can veto repeated or clearly harmful edits, but it should not replace the assigned L1 direction
- Do not repeat any exact child body already present in the history block
- Return exactly one design per slot_index listed above
- Every mutation must be structurally different from the parent and from the other mutations
\end{Verbatim}

\subsection{Attribute}
\label{app:prompts:attribute}

\begin{Verbatim}
You are classifying robot designs by their main structural archetype.

Skill library (each skill is described by its L1 archetype plus top positive sub-patterns):
{skills_block}

Designs to classify:
{designs_block}

For each design, determine which skill it structurally matches.

Use L1 condition as the PRIMARY criterion: does the design exhibit the main
load-bearing arrangement described by L1? Use top_positive_leaves as supplementary
evidence to recognize concrete instances of successful sub-patterns within
that archetype.

If the design does not clearly match any skill's L1 archetype, return null.

Constraints:
- Use structural matching only -- do not consider fitness or task performance.
- A design may match at most one skill (return the best match).
- Prefer null over a weak match. Skills that don't fit will not gain useful evidence.
- L1 dominates over L2: if a design clearly matches one skill's L1 but somewhat
  resembles another skill's positive leaves, attribute by L1.

Return JSON:
{
  "assignments": [
    {"local_index": 0, "skill_id": "skill_id" or null, "reason": "short structural reason"}
  ]
}
\end{Verbatim}

\subsection{Add}
\label{app:prompts:add}

\begin{Verbatim}
You are a robot design analyst. Your job is to add useful L1 structural skill identities for robot morphology search.

High-fitness designs (top 6 from this generation, body grids + fitness):
{high_designs}

Low-fitness designs (bottom 6 from this generation, body grids + fitness):
{low_designs}

Existing skills for this task (each with L1 condition + top 2 positive L2 leaves):
{existing_skills_block}

Look for at most ONE new L1 structural archetype in this generation.
Start from the high-fitness designs. If several share a simple main structure,
you may add it as a new skill. Use the low-fitness designs only as a light contrast
signal: it is enough if the structure is absent, broken, weaker, or less coherent there.
Be willing to add a new L1 when the high designs show a reusable structure that is not
an obvious duplicate of an existing L1 condition. Return no_add only when there is no
clear nameable structure in the high designs, or the best candidate is almost the same
as an existing L1 condition.

L1 should describe the robot's main connected load-bearing arrangement. Use simple
structure words such as frame, rail, column, bridge, arch, tripod, fork, wedge, tail,
crawler, shell, or beam. Avoid both performance goals and exact voxel-level details.
The Add step creates only the L1 identity; L2 and L3 must stay empty at birth.

Return JSON:
{
  "decision": {
    "action": "add" or "no_add",
    "inspired_obs_ids": [<int>, ...],
    "skill": {
      "skill_id": "short_name (max 3 words, snake_case, no version suffix)",
      "task_family": ["{task_name}"],
      "condition": "same text as l1.condition",
      "l1": {
        "structure": "one coarse structure word",
        "condition": "10-25 words describing one concrete structural archetype"
      },
      "l2": {
        "positive": [],
        "negative": [],
        "next_leaf_id_counter": 0
      },
      "l3": {
        "observations": [],
        "next_obs_id_counter": 0
      }
    } or null,
    "reasoning": {
      "supporting_high_labels": [<int>, ...],
      "contrast_signal": "short explanation of the high-vs-low structural difference",
      "nearest_existing_skill": "skill_id string or null",
      "duplicate_risk": "none" or "low" or "high",
      "why_add_or_no_add": "short explanation"
    }
  }
}

Rules:
- This is a lightweight discovery step, not a strict filter.
- Add when high-fitness designs reveal a simple reusable L1 structure and duplicate_risk is not high.
- Reject only generation-level summaries, performance goals, local patches, voxel-coordinate descriptions, and near-duplicate L1 conditions.
- skill_id must be max 3 words in snake_case, with no version suffix.
- If action is "add", inspired_obs_ids must list the observations that best exemplify this pattern; if no_add, set skill to null and inspired_obs_ids to [].
- For any added skill, l2.positive, l2.negative, and l3.observations MUST be empty lists.
- reasoning is audit-only. It will not be shown to later stages, so be explicit and honest.
- Candidates have gain > 0 (computed as child_fitness - parent_fitness). Only consider these for new L1 archetypes.
- When evaluating duplicate_risk, compare against both L1 conditions AND top_positive_leaves of existing skills.
  Two skills with similar archetype but different positive sub-patterns may still be distinct.
- inspired_obs_ids must list the obs_ids from the input that best exemplify this pattern (use the obs_id values shown).
{low_skill_hint}
\end{Verbatim}

\subsection{Diagnose}
\label{app:prompts:diagnose}

\begin{Verbatim}
You are diagnosing L2 leaves for a robot design skill.

Skill L1 condition: {l1_condition}

Existing L2 leaves (with current statistics):
{leaves_json}

Unassigned observations (need leaf decisions):
{unassigned_json}

Context observations (already assigned this generation; for distribution awareness only):
{context_json}

Generation statistics: gen_mean={gen_mean:.3f}, p25={gen_p25:.3f} (context only, not primary criterion).

{cold_start_note}

For each unassigned observation, decide its leaf assignment. Optionally propose new standalone leaves
that capture cross-cutting patterns visible across multiple obs.

Decision rules:
- Primary criterion: gain (= child_fitness - parent_fitness). gen_mean / p25 are weak context.
- Positive leaf creation (lenient): obs.gain > 0 AND reusable structural sub-pattern.
- Negative leaf creation (strict): obs.gain << 0 (significantly negative) AND obvious failure structure.
- Prefer "no_leaf" when there is no clear sub-pattern; the observation will be reconsidered in future generations
  (up to a hard cap of 3 attempts).
- standalone_new_leaves: only when >= 2 obs share the same not-yet-captured sub-pattern.
- description_update applies only to "match_existing" decisions.

Refinement granularity rules (apply to all claim/description text including standalone leaves):
- ALLOWED: relative regions ("lower-right corner", "upper half", "leftmost column"),
  shape language ("U-shape opening upward", "tapered top"),
  counts and proportions ("4 H_ACT", "30% RIGID", "at least 2 anchors"),
  structural relations ("anchor connects to lower rail").
- FORBIDDEN: exact voxel coordinates ("voxel at (5,4)"), numeric row/column indices
  ("row 0", "column 4"), full-body matrix templates.
- Prefer "approximately N" / "at least N" over rigid "exactly N".

Return JSON:
{
  "leaf_assignments": [
    {"obs_id": <int>, "decision": "match_existing", "leaf_id": "<id>",
      "description_update": {"mode": "overwrite" | "append" | null, "text": "..." or null}},
    {"obs_id": <int>, "decision": "new_leaf", "polarity": "positive" | "negative",
      "claim": "snake_case_1_to_4_words", "description": "structural sentence"},
    {"obs_id": <int>, "decision": "no_leaf"}
  ],
  "standalone_new_leaves": [
    {"polarity": "positive" | "negative", "claim": "...", "description": "...",
      "supporting_obs_ids": [<int>, <int>, ...]}
  ]
}
\end{Verbatim}

\subsection{Merge}
\label{app:prompts:merge}

\begin{Verbatim}
You are checking a robot design skill library for obvious duplicate L1 skill identities.

Skills in the library (L1 identity only):
{skills_full_content}

Return merge clusters only for obvious duplicate or near-duplicate L1 skill identities.
The main comparison target is L1 condition: merge skills only when they describe the
same main structural archetype in different words.

L1 structure is only a weak hint. Two skills may both use "frame", "rail", or "column"
and still be different if their L1 conditions describe different connected layouts.

Return JSON:
{
  "clusters": [
    {"group_label": "short_name (max 3 words)", "skill_ids": ["id1", "id2"], "reason": "why these are the same mechanism"}
  ]
}

Rules:
- group_label must be max 3 words, descriptive, snake_case
- Return only multi-skill clusters that should be merged; return [] if no obvious duplicates exist
- Do not create single-skill groups
- Do not merge skills just because they share the same performance goal, same task, or same broad structure word
- If there is any meaningful doubt, keep the skills separate
\end{Verbatim}

\subsection{Transfer-Context Blocks}
\label{app:prompts:transfer}

\paragraph{Same-grid transfer.}

\begin{Verbatim}
=== Transfer Context ===
The skill library below comes from a prior run on the same task
({current_env}, {source_grid} grid, source experiment "{source_exp}").
Treat the L1/L2 rules as validated patterns from previous experimentation.
Note: avg_gain values reflect the prior run's parent fitness distribution;
treat them as relative ranking signals, not absolute predictions.
\end{Verbatim}

\paragraph{Cross-grid transfer.}

\begin{Verbatim}
=== Transfer Context ===
Current task: {current_env} ({current_grid} voxel grid).
The skill library below comes from a prior run on a related task with a
{source_grid} voxel grid (source experiment "{source_exp}").

The skills' L1/L2 rules describe abstract structural principles (e.g.
"vertical rails joined by horizontal crossbeam") that should generalize
across grid sizes. The current {current_grid} grid affords richer / more
redundant structures than {source_grid}.
\end{Verbatim}

\paragraph{Elite-design addendum (same-grid).}

\begin{Verbatim}
Reference designs (top-fitness exemplars from the source run) are also
provided below. Use them as concrete examples of what the L1/L2 rules
look like in practice.
\end{Verbatim}

\paragraph{Elite-design addendum (cross-grid).}

\begin{Verbatim}
Reference designs (top-fitness exemplars) are also provided below as
concrete {source_grid} examples -- extract structural patterns from them,
do NOT copy voxel-level arrangements directly.
\end{Verbatim}

\section{Experimental Details and Hyperparameters}
\label{app:exp_details}
\label{app:exp:hyperparams}
\label{app:exp:solo_hyperparams}

\subsection{Tasks and Environment IDs}
\label{app:exp:tasks}

\begin{table}[H]
\centering
\small
\caption{\textbf{\evogym tasks and environment IDs.} We evaluate seven tasks at both $5{\times}5$ and $10{\times}10$ design scales.}
\label{tab:env_ids}
\resizebox{\linewidth}{!}{%
\begin{tabular}{llll}
\toprule
\textbf{Task} & \textbf{Objective / family} & \textbf{$5{\times}5$ env.} & \textbf{$10{\times}10$ env.} \\
\midrule
Walker        & Flat-ground locomotion        & \texttt{Walker-v0}       & \texttt{Walker-v0-10x10} \\
BridgeWalker  & Bridge traversal              & \texttt{BridgeWalker-v0} & \texttt{BridgeWalker-v0-10x10} \\
Balancer      & Balance on a beam             & \texttt{Balancer-v0}     & \texttt{Balancer-v0-10x10} \\
Carrier       & Object carrying               & \texttt{Carrier-v0}      & \texttt{Carrier-v0-10x10} \\
Climber       & Vertical traversal            & \texttt{Climber-v0}      & \texttt{Climber-v0-10x10} \\
Jumper        & Coordinated jumping           & \texttt{Jumper-v0}       & \texttt{Jumper-v0-10x10} \\
Pusher        & Object pushing                & \texttt{Pusher-v0}       & \texttt{Pusher-v0-10x10} \\
\bottomrule
\end{tabular}%
}
\end{table}

\subsection{Agent, Search, and Controller Hyperparameters}

\begin{table}[H]
\centering
\small
\caption{\textbf{Agent, search, and PPO controller hyperparameters.} Values are shared across tasks unless otherwise noted.}
\label{tab:hyperparams}
\resizebox{\linewidth}{!}{%
\begin{tabular}{lcc}
\toprule
\textbf{Hyperparameter} & \textbf{$5{\times}5$} & \textbf{$10{\times}10$} \\
\midrule
\multicolumn{3}{c}{\methodname{} agent} \\
\midrule
Designs per generation                          & $25$        & $25$ \\
\quad Path~A skill-conditioned slots             & $15$        & $15$ \\
\quad Path~B GA slots                            & $10$        & $10$ \\
Elite Pool size $k$                              & $5$         & $5$ \\
LLM mutation range (voxels)                      & $1$--$3$    & $1$--$10$ \\
LLM backbone                                     & \multicolumn{2}{c}{\texttt{gpt-5.5}} \\
LLM temperature                                  & \multicolumn{2}{c}{$1.0$} \\
Skill-weight saturation cap $\delta_{\max}$      & \multicolumn{2}{c}{$2.0$} \\
Unassigned-pool re-attribution threshold         & \multicolumn{2}{c}{$30$} \\
Prior-only generations $K$                       & --          & $5$ \\
Static elite-inject pool size                    & --          & $5$ \\
Reference design from $5{\times}5$ elite          & --          & yes in w/ref variant \\
\midrule
\multicolumn{3}{c}{Evaluation budget} \\
\midrule
Walker / BridgeWalker budget                     & $250$ evals & $1{,}000$ evals \\
Balancer / Carrier / Climber budget              & $500$ evals & $1{,}000$ evals \\
Jumper / Pusher budget                           & $750$ evals & $1{,}000$ evals \\
Seeds                         & \multicolumn{2}{c}{$0, 9, 42$} \\
\midrule
\multicolumn{3}{c}{PPO controller training} \\
\midrule
Controller architecture                           & \multicolumn{2}{c}{Multilayer perceptron} \\
Total timesteps per morphology                    & \multicolumn{2}{c}{$512{,}000$} \\
Parallel envs $n_{\text{envs}}$                   & \multicolumn{2}{c}{$4$} \\
Rollout steps $n_{\text{steps}}$                  & \multicolumn{2}{c}{$128$} \\
Update epochs                                     & \multicolumn{2}{c}{$4$} \\
Batch size                                        & \multicolumn{2}{c}{$128$} \\
Learning rate (initial)                           & \multicolumn{2}{c}{$2.5 \times 10^{-4}$} \\
Learning rate schedule                            & \multicolumn{2}{c}{linear decay to 0} \\
Clip range                                        & \multicolumn{2}{c}{$0.1$} \\
Discount $\gamma$                                 & \multicolumn{2}{c}{$0.99$} \\
GAE $\lambda$                                     & \multicolumn{2}{c}{$0.95$} \\
Entropy coefficient                               & \multicolumn{2}{c}{$0.01$} \\
Value-loss coefficient                            & \multicolumn{2}{c}{$0.5$} \\
Max gradient norm                                 & \multicolumn{2}{c}{$0.5$} \\
Log / eval interval (updates)                     & \multicolumn{2}{c}{$50$} \\
\bottomrule
\end{tabular}%
}
\end{table}

\subsection{Validity Checks}
\label{app:validity}

Every proposed body is checked before simulation. A valid body must contain only legal robot-body voxel values, include at least one actuator, and form a connected component after empty cells are ignored. The environment-reserved \texttt{FIXED} voxel type is not allowed inside the robot body. Invalid LLM outputs are repaired only when the edit is local and unambiguous; otherwise, the proposal slot is replaced by GA mutation. These checks ensure that all reported comparisons spend their simulation budget on feasible morphology evaluations rather than on malformed bodies.

\section{Additional Results and Qualitative Analysis}
\label{app:additional_results}

\subsection{Representative Skill Types}
\label{app:skill_examples}

The learned Walker skill library is visualized in the main paper (Figure~\ref{fig:walker_library}). Here we summarize additional skill types that recur across tasks and help explain why the library transfers better than raw upsampling.

\paragraph{Load-bearing frame.}
A frame skill describes rigid or semi-rigid support around actuators so that actuation produces task-relevant motion rather than body collapse. This skill is useful for Walker, BridgeWalker, and Carrier because each task rewards controlled force transmission through the body.

\paragraph{Compliant contact pad.}
A contact skill places soft material near the interaction surface while preserving stiffer internal support. It is useful in Carrier and Pusher because the body must maintain contact with an object during motion rather than merely moving itself.

\paragraph{Actuator anchor.}
An actuator-anchor skill pairs active voxels with neighboring rigid or soft support. The rule prevents isolated actuation from producing local deformation, rotation, or internal cancellation without useful displacement.

\subsection{Component Ablations}
\label{app:ablations}

To probe which mechanisms of \S\ref{sec:method} carry the gain, we disable one component of the library pipeline at a time under the same $5{\times}5$ cold-start protocol as Table~\ref{tab:exp2_5x5}, on Walker and Pusher, mean of three seeds ($0$, $9$, $42$). (A1) \textbf{No Diagnose} skips the \textsc{Diagnose} call, so $L_2$ rules never form or update. (A2) \textbf{No Merge} skips the per-generation \textsc{Merge} pass, letting $L_1$ duplicates accumulate. (A3) \textbf{Pure LLM} drops GA mutation so all $25$ slots come from skill-conditioned LLM mutation. (A4) \textbf{No $L_2$+$L_3$} strips $L_2$ rules and $L_3$ observations from the \textsc{Propose} prompt.

\begin{figure}[t!]
  \centering
  \includegraphics[width=\columnwidth]{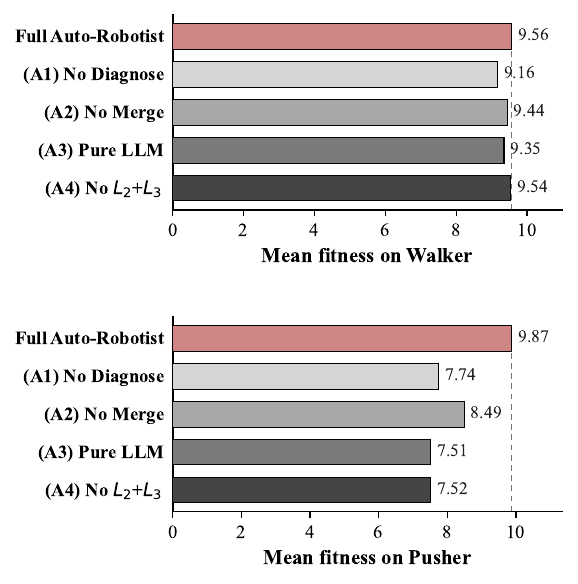}
  \caption{Best fitness on Walker and Pusher per ablation condition, averaged over three seeds.}
  \label{fig:ablation_multiseed}
\end{figure}

On Pusher (Figure~\ref{fig:ablation_multiseed}), (A3) at $-24.0\%$ indicates that GA mutation remains a source of structural exploration the skill-conditioned proposer alone cannot recover. (A4) at $-23.8\%$ indicates that $L_2$ rules sharpen the abstract $L_1$ archetypes into actionable guidance, and $L_3$ observations ground both in concrete voxel instances the LLM can interpret. (A1) at $-21.6\%$ and (A2) at $-14.0\%$ further confirm that library maintenance itself is also necessary on this task. All four mechanisms thus contribute materially to the gain on Pusher.

On Walker, the same four ablations cost only $0.2$--$4.3\%$, an order of magnitude less than on Pusher, because every condition saturates near the same ceiling regardless of which component is removed. The pattern matches \S\ref{sec:experiments}: each component matters most where the task leaves more room for the search to go wrong.

\subsection{LLM Cost}
\label{app:llm_cost}

For every API call, we tracked the issuing stage, prompt and completion token counts, and latency. Table~\ref{tab:llm_cost}
  reports the aggregate usage for the $10{\times}10$ warm-start \emph{w/ ref} setting, which is the most expensive
  configuration evaluated in this paper. This setting runs for $40$ generations and performs $1{,}000$ morphology evaluations
  using \texttt{gpt-5.5}, priced at \$5 per million input tokens and \$30 per million output tokens.

\begin{table}[H]
\centering\footnotesize
\setlength{\tabcolsep}{5pt}
\begin{tabular}{lcccc}
\toprule
\textbf{Task} & Calls & In (M) & Out (M) & Cost (\$) \\
\midrule
Walker        & $349$ & $2.27$ & $1.35$ & $52.0$ \\
BridgeWalker  & $389$ & $2.06$ & $1.52$ & $55.8$ \\
Balancer      & $386$ & $3.06$ & $1.57$ & $62.3$ \\
Carrier       & $346$ & $2.48$ & $1.29$ & $51.1$ \\
Climber       & $385$ & $2.85$ & $1.42$ & $56.8$ \\
Jumper        & $344$ & $2.49$ & $1.41$ & $54.8$ \\
Pusher        & $376$ & $2.36$ & $1.42$ & $54.3$ \\
\bottomrule
\end{tabular}
\caption{Measured LLM usage per $10{\times}10$ warm-start \emph{w/ ref} run}
\label{tab:llm_cost}
\end{table}

\paragraph{Cost analysis.} LLM usage is stable across tasks, and none of the logged calls returned an API error. Most of the wall-clock time goes to PPO training rather than the LLM, and only a small fraction of skill-conditioned proposals ever need repair. This overhead is modest compared to what it buys in search quality under the same evaluation budget.


\section{Extended Discussion}
\label{app:discussion}

\subsection{What Makes a Design Skill Transferable?}
\label{sec:disc:transferable}

The experiments suggest that transferable design skills sit between two extremes. A raw elite body is too concrete: it may work at one grid size but fail when naively resized. A vague reflection such as ``be stable'' is too abstract: it does not constrain a proposal enough to improve search. Useful skills name structural relations that remain meaningful under scale changes, such as a support frame around actuators, a compliant contact surface backed by rigid structure, or an actuator band anchored by passive material. These relations are the right level of abstraction because they can be instantiated differently while preserving function.

\subsection{Why Evidence Grounding Matters}
\label{sec:disc:evidence}

Natural language is expressive enough to describe physical design principles, but it can also produce plausible explanations unsupported by evidence. \methodname addresses this risk by treating a skill as an evidence-linked record rather than a free-form reflection. $L_3$ observations preserve the evaluated bodies, tasks, scales, and gains; $L_2$ rules summarize repeated positive or negative patterns; and the sampling weight in Equation~\ref{eq:skill_weight} gives stronger influence to rules with consistent support. This design makes the library auditable: a user can inspect which evaluations caused a rule to matter, and future updates can revise or merge rules when new evidence contradicts them.

\subsection{Boundary Conditions for Transfer}
\label{sec:disc:failures}

Transfer is strongest when source-scale skills describe relations that remain valid at the target scale. Carrier is the clearest example: load-bearing and contact-stabilizing motifs learned at $5{\times}5$ remain useful at $10{\times}10$. Some tasks additionally require geometric anchoring. BridgeWalker, Balancer, and Climber without a reference design illustrate this case: the library alone captures useful structural priors, but the reference-conditioned variant supplies topology and environment-specific anchoring that abstract rules do not fully determine.

These cases highlight two design pressures for future memory systems. First, useful rules may interfere when they prescribe incompatible edits, such as placing a rigid spine where another rule prefers a central actuator band. Second, a library must avoid crystallizing a concept before enough evidence distinguishes a general principle from a lucky design. \methodname addresses these pressures through evidence tracking and merging; richer uncertainty-aware retrieval and explicit rule-conflict resolution are natural next steps.


\section{Responsible NLP, Artifacts, and Reproducibility}
\label{app:responsible_nlp}

This appendix summarizes the responsible-research considerations for \methodname{}. The work studies simulated robot morphology search, not human-subject data or user-facing deployment. The main responsibility issues are therefore artifact use, reproducibility, simulator-to-real limitations, and auditability of LLM-assisted design decisions.

\subsection{Scope and Artifacts}
\label{app:responsible_scope}

\paragraph{Scope.}
\methodname{} is a research system for studying whether language agents can convert expensive morphology-search evaluations into reusable design knowledge. The method operates in \evogym{} and uses simulated robot morphologies, controller-training results, fitness values, LLM proposal traces, and learned skill-library records. It does not collect, annotate, infer, or release human-subject data. No personally identifying information, demographic attributes, private text, or user-generated natural-language corpus is used.

\paragraph{Scientific artifacts.}
The main scientific artifacts used in this work are \evogym{} environments, PPO controller training, the genetic-algorithm baseline, LLM prompts, generated voxel morphologies, simulation logs, and learned skill libraries. The paper cites \evogym{} and PPO in the main text, documents prompt templates in Appendix~\ref{app:prompts}, and reports task descriptions, validity constraints, evaluation budgets, metrics, and hyperparameters in Appendix~\ref{app:exp_details}. The artifacts created by this work are intended for research and reproducibility: they should be used to study morphology search, language-agent memory, and simulator-grounded design rules, not as deployment-ready hardware designs.

\paragraph{Licenses and intended use.}
\evogym{} is a public research benchmark for evolving soft robots. Our use is consistent with this intended purpose: we compare morphology-search algorithms in simulation and do not deploy generated designs as physical robots. Any released code, prompt files, robot-body grids, logs, or skill-library records will preserve attribution to the original benchmark and third-party packages. Released artifacts should include explicit license information and should be treated as research artifacts rather than safety-certified robot designs.

\paragraph{Computational experiments and budget.}
All experiments are computational. We report compute in platform-independent units because wall-clock time depends on parallelization and hardware. Each morphology evaluation trains a PPO controller for $5.12{\times}10^5$ environment steps. At the $5{\times}5$ scale, Walker and BridgeWalker use 250 morphology evaluations; Balancer, Carrier, and Climber use 500; and Jumper and Pusher use 750. At the $10{\times}10$ scale, all tasks use 1,000 morphology evaluations. \methodname{} evaluates 25 designs per generation, with 15 skill-conditioned LLM-proposal slots and 10 GA-mutation slots. The LLM is used as a fixed prompting component for \textsc{Propose}, \textsc{Add}, \textsc{Diagnose}, and \textsc{Merge}; its parameters are not trained or fine-tuned.

\paragraph{Statistical reporting.}
The paper reports best-fitness-so-far curves, endpoint fitness, convergence speedup, and lead fraction under a fixed evaluation protocol. These quantities are descriptive comparisons of search behavior under equal morphology-evaluation budgets. Unless otherwise stated, the reported values should not be interpreted as formal statistical-significance claims. Future work should add multi-seed confidence intervals and wall-clock/hardware-normalized cost analysis.

\paragraph{Risks and safety.}
The main risks are not privacy or fairness risks, because the work uses no personal or demographic data. The relevant responsibility risks concern robotics and automation. A language agent may discover morphologies that exploit simulator artifacts, overfit to benchmark-specific physics, or fail under real-world manufacturing tolerances. We therefore frame \methodname{} as a simulator-facing research prototype. Any physical deployment should require human engineering review, stress testing, uncertainty-aware validation, and safety constraints outside the training simulator.

\paragraph{Auditability.}
The proposed skill library is designed to make LLM-assisted design more auditable. Each skill stores an archetype, positive and negative rules, and supporting observations. Thus, a generated proposal can be traced to the retrieved skill, the rule that motivated the edit, and the simulation evidence behind that rule. This evidence link is important because natural language alone can produce plausible but unsupported explanations.

\paragraph{AI-assistant use.}
This work uses LLMs as part of the proposed method: the agent prompts a fixed LLM for proposal, skill addition, diagnosis, and merging. These prompts are documented in Appendix~\ref{app:prompts}. Separately, AI assistants were used for drafting, editing, formatting, and project-organization support. They are not authors, did not determine the scientific claims, and do not replace author responsibility. The authors remain responsible for the correctness of experiments, citations, figures, tables, claims, and final submission content.

\subsection{Social Impact}
\label{app:social_impact}

\methodname{} is intended as a research system for studying how language agents can preserve useful knowledge from expensive scientific evaluations. Its potential benefits are methodological: making morphology search more sample-efficient, making design priors inspectable through evidence-linked skills, and reducing repeated simulation effort by reusing what earlier searches discovered.

The main risks arise if such agents are moved from simulation to physical design without safeguards. Automatically generated morphologies can exploit simulator artifacts, fail under real-world tolerances, or encode brittle design heuristics that appear reliable only under a narrow benchmark. We therefore frame the system as a simulator-facing research prototype, not as a direct hardware-deployment tool. Any physical use should require human engineering review, safety constraints, stress testing, uncertainty-aware validation, and validation outside the training simulator.

Because the work uses no personal or demographic data, direct privacy and fairness risks are limited. The more relevant responsibility concern is auditability. The proposed library stores not only skill names but also rules, supporting observations, and failure evidence, so users can inspect why a design prior was retrieved rather than treating the LLM proposal as an opaque recommendation.

\section{Additional Related Work}
\paragraph{Curriculum, transfer, and abstraction in design search.}
A second relevant thread studies how agents acquire reusable structure across tasks. Curriculum learning and automatic curriculum learning shape the order of experience to improve generalization~\citep{bengio2009curriculum,portelas2020automatic}, while environment-design methods seek training distributions that induce robust transfer~\citep{dennis2020emergent}. Modular meta-learning and program-library induction show that reusable components can make future learning more efficient~\citep{alet2018modular,ellis2023dreamcoder}. \methodname shares this motivation but operates in a different setting: it does not pretrain a model or learn a latent module. Instead, it writes simulator evidence into natural-language design rules that can be inspected, edited, and reused by an LLM agent.
\end{document}